\documentclass[sigconf,nonacm,screen]{acmart}

\usepackage{enumitem}

\AtBeginDocument{%
  }

\setcopyright{acmlicensed}
\copyrightyear{2018}
\acmYear{2018}
\acmDOI{XXXXXXX.XXXXXXX}

\acmConference[Conference acronym 'XX]{Make sure to enter the correct
  conference title from your rights confirmation emai}{June 03--05,
  2018}{Woodstock, NY}

\acmISBN{978-1-4503-XXXX-X/18/06}

\acmSubmissionID{829}

\begin{document}

\title{LAM: Large Avatar Model for One-shot Animatable Gaussian Head}

\author{Yisheng He}
\email{heyisheng.hys@alibaba-inc.com}
\authornote{Equal Contribution}
\affiliation{%
  \institution{Tongyi Lab, Alibaba Group}
  \country{China}
}

\author{Xiaodong Gu}
\authornotemark[1]
\email{dadong.gxd@alibaba-inc.com}
% \orcid{0000-0003-2623-7973}
\affiliation{%
  \institution{Tongyi Lab, Alibaba Group}
  \country{China}
}

\author{Xiaodan Ye}
\email{doris.yxd@alibaba-inc.com}
\affiliation{%
  \institution{Tongyi Lab, Alibaba Group}
  \country{China}
}
\author{Chao Xu}
\email{eric.xc@alibaba-inc.com}
\affiliation{%
  \institution{Tongyi Lab, Alibaba Group}
  \country{China}
}
\author{Zhengyi Zhao}
\email{bushe.zzy@alibaba-inc.com}
\affiliation{%
  \institution{Tongyi Lab, Alibaba Group}
  \country{China}
}
\author{Yuan Dong}
\email{dy283090@alibaba-inc.com}
\authornote{Corresponding Author}
\affiliation{%
  \institution{Tongyi Lab, Alibaba Group}
  \country{China}
}
\author{Weihao Yuan}
\email{qianmu.ywh@alibaba-inc.com}
\authornotemark[2]
\affiliation{%
  \institution{Tongyi Lab, Alibaba Group}
  \country{China}
}
\author{Zilong Dong}
\email{list.dzl@alibaba-inc.com}
\affiliation{%
  \institution{Tongyi Lab, Alibaba Group}
  \country{China}
}
\author{Liefeng Bo}
\email{liefeng.bo@alibaba-inc.com}
\affiliation{%
  \institution{Tongyi Lab, Alibaba Group}
  \country{China}
}

\renewcommand{\shortauthors}{He et al.}

\begin{abstract}
We present LAM, an innovative \textbf{L}arge \textbf{A}vatar \textbf{M}odel for animatable Gaussian head reconstruction from a single image. Unlike previous methods that require extensive training on captured video sequences or rely on auxiliary neural networks for animation and rendering during inference, our approach generates Gaussian heads that are immediately animatable and renderable. Specifically, LAM creates an animatable Gaussian head in a single forward pass, enabling reenactment and rendering without additional networks or post-processing steps. This capability allows for seamless integration into existing rendering pipelines, ensuring real-time animation and rendering across a wide range of platforms, including mobile phones. The centerpiece of our framework is the canonical Gaussian attributes generator, which utilizes FLAME canonical points as queries. These points interact with multi-scale image features through a Transformer to accurately predict Gaussian attributes in the canonical space. The reconstructed canonical Gaussian avatar can then be animated utilizing standard linear blend skinning (LBS) with corrective blendshapes as the FLAME model did and rendered in real-time on various platforms. Our experimental results demonstrate that LAM outperforms state-of-the-art methods on existing benchmarks. Our code and video are available at~\url{https://aigc3d.github.io/projects/LAM/}.
\end{abstract}

\begin{CCSXML}
<ccs2012>
   <concept>
       <concept_id>10010147.10010178.10010224.10010245.10010254</concept_id>
       <concept_desc>Computing methodologies~Reconstruction</concept_desc>
       <concept_significance>500</concept_significance>
       </concept>
   <concept>
       <concept_id>10010147.10010178.10010224.10010240.10010242</concept_id>
       <concept_desc>Computing methodologies~Shape representations</concept_desc>
       <concept_significance>500</concept_significance>
       </concept>
   <concept>
       <concept_id>10010147.10010178.10010224.10010240.10010243</concept_id>
       <concept_desc>Computing methodologies~Appearance and texture representations</concept_desc>
       <concept_significance>500</concept_significance>
       </concept>
 </ccs2012>
\end{CCSXML}

\ccsdesc[500]{Computing methodologies~Reconstruction}
\ccsdesc[500]{Computing methodologies~Shape representations}
\ccsdesc[500]{Computing methodologies~Appearance and texture representations}

\keywords{Animatable Head Avatar, Gaussian Splatting, Single-shot Reconstruction, Real-time Animation and Rendering}

\begin{teaserfigure}
  \centering
    \includegraphics[width=1.\linewidth]{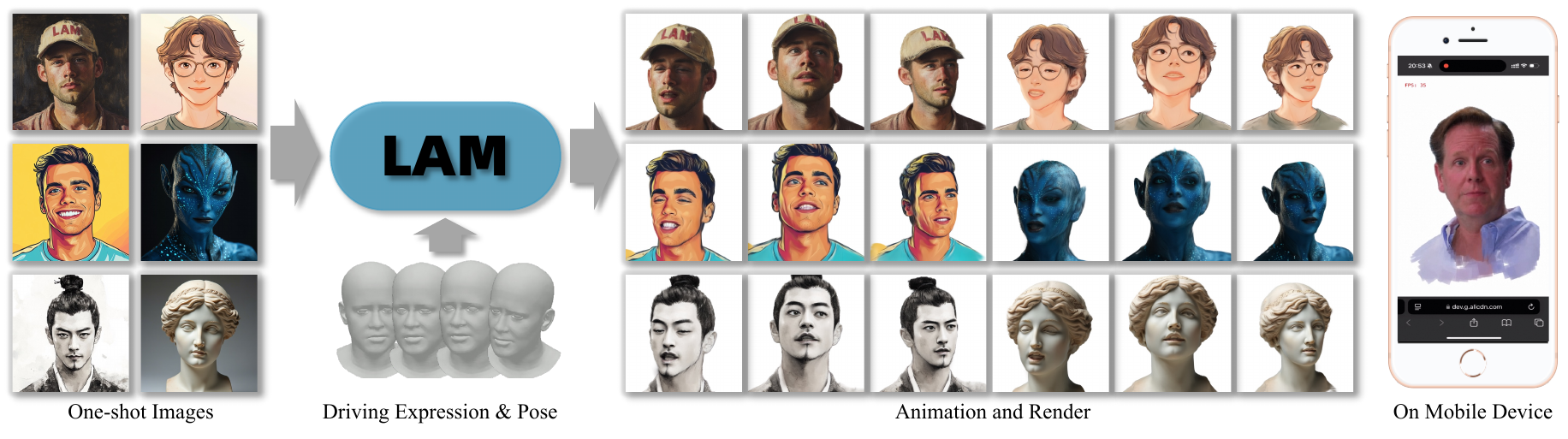}
  \caption{LAM creates animatable Gaussian heads with one-shot images in a single forward pass. The reconstructed Gaussian avatar can be reenacted and rendered on various platforms in real time.}
  \Description{}
  \label{teaser}
\end{teaserfigure}

\maketitle

\section{Introduction}

The reconstruction and animation of head avatars from a single image is an important area of research within computer graphics and computer vision. This area has a wide range of practical applications, including 
online meetings, filmmaking, the gaming industry, and virtual reality.
Researchers have tackled this challenge using different methodologies, primarily focusing on 2D and 3D generative models.

Early 2D methods~\cite{DBLP:conf/cvpr/KarrasLAHLA20,DBLP:conf/cvpr/IsolaZZE17,DBLP:conf/nips/GoodfellowPMXWOCB14} harness the capabilities of convolutional neural networks and generative adversarial networks to depict facial expressions and poses via warping fields applied to source images. Benefiting from advances in image and video diffusion networks, more recent 2D-based works~\cite{DBLP:journals/corr/abs-2410-07718,DBLP:journals/corr/abs-2406-08801,DBLP:conf/eccv/TianWZB24} get improved results with diffusion techniques. 
Although achieving impressive results, 2D solutions still face challenges related to long generation times and significant computational resource demands.
Moreover, 2D techniques often struggle with extreme variations in poses and expressions due to their lack of explicit three-dimensional structures.

On the other hand, the emergence of advanced 3D synthesis techniques, such as Neural Radiance Fields (NeRFs) and Gaussian Splatting, has enhanced the consistency of multi-view avatar reconstruction and animation. Nonetheless, NeRF-based~\cite{DBLP:conf/cvpr/GafniTZN21,DBLP:conf/eccv/KiMC24,DBLP:conf/cvpr/BaiFWZSYS23,PointAvatar,Nerfies,DBLP:conf/siggraph/YuFZWYBCSWSW23,DBLP:conf/cvpr/MaZQLZ23,DBLP:conf/cvpr/LiZWZ0CZWB023} methods usually require extensive optimization from a single image for each individual before effective use, and their slow rendering speeds limit their application in scenarios demanding real-time performance. 

Recent developments, like GAGAvatar~\cite{GAGAvatar}, introduced a framework for single-shot Gaussian avatar generation and animation that capitalizes on the quick rendering prowess of Gaussian Splatting. 
However, these methods usually rely heavily on complex 2D neural post-processing to achieve optimal outcomes, thus they are not pure 3D solutions and may encounter similar problems presented in 2D solutions.
In addition, their severe reliance on neural network post-processing makes their integration into traditional rendering pipelines across various platforms cumbersome.

In this paper, we introduce a large avatar model for generating animatable Gaussian head avatars. After one pass forward of the network given a single image, our approach generates the Gaussian avatar that can be instantly reenacted and rendered without any post-processing network. This capability allows for seamless integration into existing rendering pipelines and ensures efficient performance across a wide range of devices, including mobile devices. 

However, this task presents certain challenges. 
\textbf{1)} The first hurdle lies in enabling the animation of the reconstructed Gaussian without relying on extra neural networks. Unlike previous studies that use neural networks~\cite{GAGAvatar} for animation and rendering, our method generate Gaussian avatar based on the FLAME animation model, which operates and renders without additional neural networks, allowing real-time application on various platforms.
\textbf{2)} The second challenge involves reconstructing an animatable Gaussian from a single image. To overcome this, we propose three main designs for our framework. 
\textbf{i)} Firstly, unlike previous methods that reconstruct shapes from implicit tri-plane~\cite{LGM} or image plane~\cite{GAGAvatar}, we represent the shape with explicite point cloud initialized from canonical FLAME vertices. In this way, we leverage the prior avatar shape resides in the FLAME model to alleviates the reconstruction challenge. 
\textbf{ii)} Secondly, instead of directly reconstructing the avatars with various poses and expressions, we reconstruct all Gaussian avatars in the canonical space with the same expression and pose. Such a unified design not only enables convenient animation during inference but also mitigates the reconstruction complexities by reducing the shape and pose variety. 
\textbf{iii)} Finally, unlike previous methods~\cite{transhuman} that only utilize the painted image features for texture and shape reasoning, our framework thoroughly reason on both local and global image features within a cross-attention mechanism, improving the reconstruction quality and texture fidelity. 

The principal contributions of this work can be summarized as follows:

\begin{itemize}[leftmargin=*]
\item We introduce a large avatar model for generating the animatable Gaussian head avatar from one image, allowing instant animation and rendering without additional post-processing.
\item The generated Gaussian avatars from our framework can be seamlessly integrated into traditional rendering pipelines, supporting real-time animation and rendering across various platforms.
\item Built upon our large animatable Gaussian avatar model, we enable the efficient generation and stylization of animatable Gaussian avatars from a single text prompt or image.
\item Experimental results on benchmark datasets demonstrate the efficacy and efficiency of our approach.
\end{itemize}

\section{Related work}

Recent advances in single-image animatable head avatar generation can be categorized into mainly 2D-based and 3D-based approaches. 

\paragraph{\bf Image to 2D Animatable Avatar.}
2D-based methods, leveraging the power of convolutional neural networks (CNNs)~\cite{DBLP:conf/cvpr/KarrasLAHLA20,DBLP:conf/cvpr/IsolaZZE17,DBLP:conf/nips/GoodfellowPMXWOCB14}, often employ generative adversarial networks (GANs)~\cite{DBLP:conf/cvpr/StyleGAN} for direct image synthesis. Early approaches~\cite{DBLP:conf/cvpr/WangDYSW23,DBLP:conf/cvpr/BurkovPGL20,DBLP:conf/iccv/ZakharovSBL19} focus on injecting expression and pose features into the generator network, often utilizing architectures like U-Net or StyleGAN~\cite{DBLP:conf/cvpr/StyleGAN}.
Some other 2D methods~\cite{DBLP:journals/corr/abs-2407-03168,DBLP:conf/cvpr/ZhangQZZW0CW023,DBLP:conf/cvpr/HongZS022,DBLP:conf/mm/DrobyshevCKILZ22,DBLP:conf/cvpr/BurkovPGL20,DBLP:conf/nips/SiarohinLT0S19} represent expressions and poses as warping fields applied to the source image. 
Benefiting from advances in image and video diffusion networks, more recent 2D-based works~\cite{DBLP:journals/corr/abs-2410-07718,DBLP:journals/corr/abs-2406-08801,DBLP:conf/eccv/TianWZB24} get improved results with diffusion techniques. 
However, these methods still face challenges related to long generation times and significant computational resource demands. Audio-driven 2D control methods~\cite{DBLP:conf/cvpr/ZhangCWZSGSW23,DBLP:journals/corr/abs-2211-12368,DBLP:conf/iccv/GuoCLLBZ21} are easy to use but cannot explicitly control facial expressions and poses. 2D-based techniques often struggle with large pose or expression variations due to the lack of an explicit 3D structure, sometimes producing unrealistic distortions or identity changes. While some 2D methods~\cite{SadTalker,StyleHEAT,Pirenderer,DBLP:conf/cvpr/WangM021,MegaPortraits} incorporate 3D Morphable Models (3DMMs)~\cite{DBLP:conf/fgr/GerigMBELSV18,DBLP:journals/tog/LiBBL017,DBLP:conf/avss/PaysanKARV09,DBLP:conf/siggraph/BlanzV99} to mitigate these issues, they typically cannot achieve free-viewpoint rendering. 

\vspace{-0.1in}

\begin{figure*}[h]
    \centering
    \includegraphics[width=0.9\linewidth]{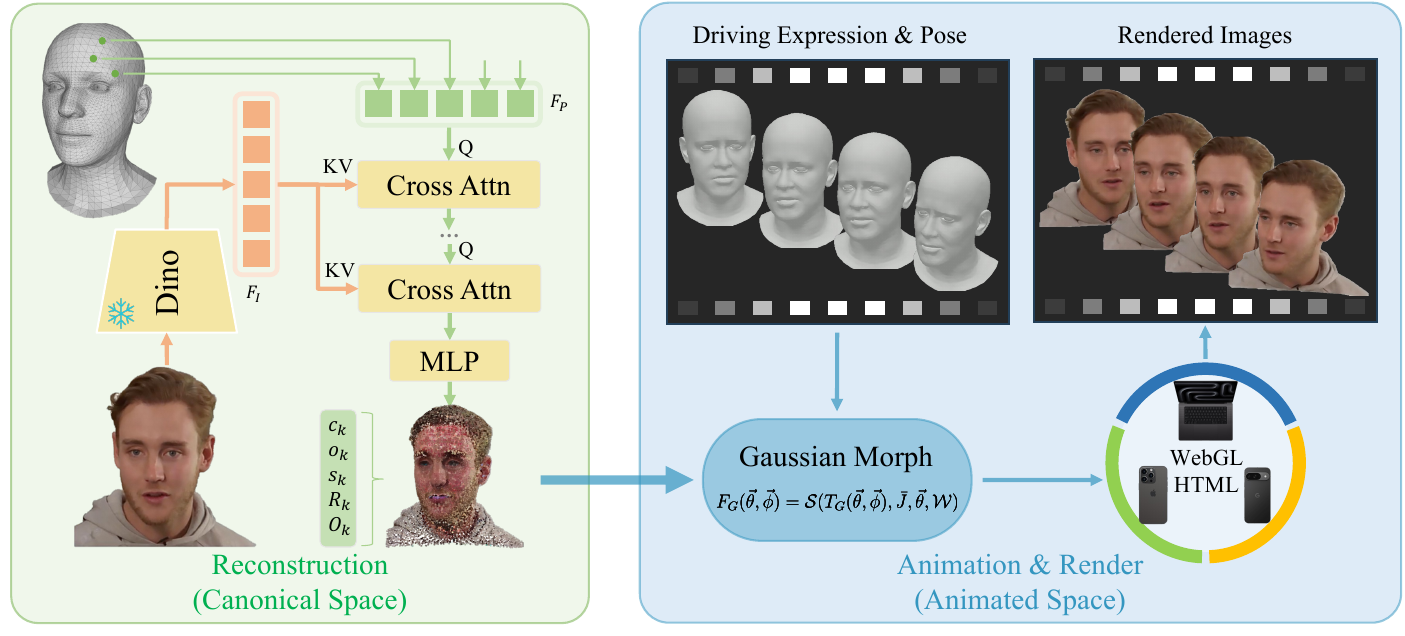}
    \caption{\textbf{Overall Framework.} Our framework utilizes learnable query features attached to FLAME vertices to perform cross-attention with the extracted multi-level image features. The extracted features are then decoded to reconstruct the Gaussian avatar in the canonical space, which can be animated utilizing standard linear blend skinning (LBS) and corrective blendshapes as the FLAME model did and rendered in real-time on various platforms.}
    \label{fig:framework}
\end{figure*}

\paragraph{\bf Image to 3D Animatable Avatar.}
3D-aware methods offer improved geometric consistency and free-viewpoint rendering capabilities. Early 3D approaches~\cite{DBLP:conf/eccv/KhakhulinSLZ22,DBLP:conf/cvpr/XuYCWDJT20} utilize 3DMMs for head avatar reconstruction. With the advent of Neural Radiance Fields (NeRFs)~\cite{DBLP:conf/eccv/MildenhallSTBRN20}, many recent methods~\cite{DBLP:conf/siggraph/YuFZWYBCSWSW23,DBLP:conf/cvpr/MaZQLZ23,DBLP:conf/cvpr/LiZWZ0CZWB023,GPAvatar,ye2024real3d,deng2024portrait4d,deng2024portrait4d2,DBLP:conf/eccv/KiMC24,DBLP:conf/cvpr/BaiFWZSYS23,PointAvatar,Nerfies,INSTA} have adopted this representation for higher fidelity, particularly in modeling fine details like hair. However, NeRF-based~\cite{DBLP:conf/cvpr/ZhangZLHLWGCL024,HAvatar,DBLP:conf/cvpr/BaiTHSTQMDDOPTB23,AD-NeRF,DBLP:journals/tog/GaoZXHGZ22,DBLP:journals/tog/ParkSHBBGMS21,DBLP:conf/cvpr/AtharXSSS22,DBLP:journals/corr/abs-2112-05637,DBLP:conf/iccv/TretschkTGZLT21,DBLP:conf/cvpr/GafniTZN21,DBLP:conf/eccv/KiMC24,DBLP:conf/cvpr/BaiFWZSYS23,PointAvatar,Nerfies,DBLP:conf/siggraph/YuFZWYBCSWSW23,DBLP:conf/cvpr/MaZQLZ23,DBLP:conf/cvpr/LiZWZ0CZWB023} approaches often require extensive training data, including multi-view or single-view videos, raising privacy concerns and limiting generalization to unseen identities. Some methods~\cite{DBLP:conf/cvpr/SunWWLZZL23,DBLP:conf/3dim/ZhuangMKS22,DBLP:journals/pami/SunWZHWL24,DBLP:journals/tvcg/TangZYZCMW24,DBLP:conf/iclr/XuZLZBFS23} bypass this data requirement by training generators with random noise and then inverting them for identity-specific reconstruction, but inversion accuracy remains a challenge. Test-time optimization offers another alternative, but its computational cost limits practical applications. Several recent works~\cite{goha2023,hidenerf2023,gpavatar2024,ye2024real3d,ma2024cvthead,deng2024portrait4d,deng2024portrait4d2,GGHead} have explored one-shot 3D head reconstruction to address the limitations of data requirements and computational cost. These methods employ various techniques, such as tri-plane features, deformation fields, point-based expression fields, and vertex-feature transformers. Despite these advancements, NeRF-based methods often struggle with real-time rendering. 
Recently, 3D Gaussian Splatting~\cite{GaussianSplatting} has emerged as a promising alternative, offering both high-quality results and fast rendering speeds. However, existing Gaussian Splatting methods~\cite{GaussianAvatar,DBLP:conf/cvpr/XuCL00ZL24} typically rely on video data for training for each person, limiting their ability to generalize to new identities. Instead, the most recent work, GAGAvatar~\cite{GAGAvatar}, proposes a one-shot 3D Gaussian-based head avatar generation method. However, it still relies heavily on complex 2D neural post-processing to achieve optimal animation outcomes, thus it is not a pure 3D solution and the extra neural network hinders its application on various platforms. In contrast, our work generates Gaussian heads that are immediately animatable and renderable without additional networks or post-processing steps, enabling seamless integration into existing rendering pipelines for real-time animation and rendering across a wide range of platforms, including mobile phones. 
\section{Methodology}

\subsection{Overview}
Our approach begins with the development of a canonical Gaussian attribute generator, utilizing a Transformer network. We incorporate pre-trained Vision Transformer (ViT) weights to effectively extract features at multiple levels from one single-view image. To fully leverage the prior shape information and expression coefficients provided by the FLAME~\cite{DBLP:journals/tog/LiBBL017} model, we employ vertices from the FLAME model as initial Gaussian positions. These vertices reason geometric relationships through stacked self-attention and extract corresponding features from the images within stacked cross-attention modules, allowing us to predict the canonical Gaussian attributes and deformation offsets to improve the shape accurately.
Given a new expression and pose, we utilize the animation weights from the FLAME model to animate the reconstructed canonical Gaussian points accordingly. This process facilitates the rendering of the image, reflecting the desired expression and pose with accuracy and enabling interactive frame rates of animation and rendering on a wide range of computing platforms, including mobile phones.

\vspace{-0.05in}

\subsection{Canonical Gaussian Head Avatar Generation}
\label{sec:CanonicalGSGen}
In this section, we outline our approach for reconstructing an animatable avatar in the canonical space from a single input image. Our objective is to create an avatar that can be animated and rendered seamlessly, without the need for additional neural networks or post-processing. This allows for straightforward integration into existing rendering pipelines and enables interactive frame rates across various computing devices.

\paragraph{\textbf{Canonical Gaussian Head Avatar Reconstruction.}} We draw inspiration from the success of Gaussian Splatting~\cite{GaussianSplatting} in novel view synthesis and its efficient rendering capabilities. Therefore, we design our model to generate the Gaussian head avatar. However, this task presents certain challenges. The first hurdle lies in enabling the animation of the reconstructed Gaussian without relying on extra neural networks. Previous studies~\cite{DBLP:conf/cvpr/XuCL00ZL24} have explored dynamic Gaussians that utilize multi-layer perceptrons (MLPs) to transform canonical Gaussians into new expressions by predicting the residual attributes and rendered images with new expressions. However, the reliance on additional neural networks hampers compatibility with traditional rendering workflows and hinders real-time rendering on platforms with less compatibility, such as mobile phones. To address this issue, we incorporate insights from the animatable FLAME model, which is animatable without extra neural networks. By using FLAME as a foundation for constructing our animatable Gaussian, we leverage its animation pipeline to animate our reconstructed avatar.

The second challenge involves reconstructing an animatable Gaussian from a single image. To overcome this, we propose three main designs for our framework: 
1) Firstly, unlike previous methods~\cite{LGM,OTAvatar} that represent and reconstruct shapes from tri-plane representation, we represent the shape with a point-cloud-like technique and initialize the point position with canonical vertices of the FLAME model. In this way, we can build the avatar with explicit shape and leverage the shape prior that resides in the FLAME model, which alleviates the reconstruction challenge. 
2) Secondly, instead of directly reconstructing the avatars with various poses and expressions, we reconstruct all Gaussian avatars within the same canonical coordinate with the same canonical expression and pose, which are then animated to the target pose and expression. Such a unified design not only enables convenient animation during inference but also mitigates the complexities associated with the reconstruction by reducing the shape and pose variety. 
3) Finally, unlike previous methods~\cite{transhuman} that only utilize the painted image features lifted from projections for texture and shape reasoning, we thoroughly utilize both local and global features extracted from the source image by building cross-attention modules between the point and image features, which improve the reconstruction quality and texture fidelity.

As illustrated in Fig. \ref{fig:framework}, we begin by employing the pre-trained DinoV2 model~\cite{DinoV2} to extract features from the input image. By fusing features derived from both shallow and deep layers, we obtain image features $F_{I}$, that capture essential local details, such as hair strands and wrinkles, while ensuring global resemblance to the original image. To incorporate prior shape information from the FLAME model, we initial the Gaussian positions based on the FLAME's vertices after shape blendshapes. Then we attach learnable features to each point as learnable query features $F_{P}$. In implementation, we utilize multi-layer perceptron (MLP) to project the positional encoding~\cite{NeRF} of each point to get $F_{P}$. These learnable features are trained to extract features from the images feature $F_I$ with stacked cross-attention modules $\mathcal{C} = \{\mathcal{C}_i\}^{L}_{i=1}$. Given the query point features $F_{P}$ and the extracted multi-level image features $F_{I}$, we flatten the features and process them with stacked cross-attention modules as:
\begin{equation}
    F_{P_{i}} = \mathcal{C}_{i}(F_{P_{i-1}}, F_I),
\end{equation}
where $\mathcal{C}_i$ is the $i_{th}$ cross-attention module, $F_{P_{i}}$ the $i_{th}$ layer's queried features given previous layers' output $F_{P_{i-1}}$ as query, with $F_{P_{1}} = F_P$.
This architecture enables the model to interpret geometric information and relationships within the point cloud via self-attention, while simultaneously using cross-attention to extract relevant appearance features from the image. Another advantage of this framework design is that we can leverage the scaling law of the Transformer architecture, which has been proven on ChatGPT~\cite{} for text generation and ~\cite{ScalableDiffusionModelsWTransformers} for image generation. 
The advantage of this framework design is that we can leverage the scaling law of the Transformer architecture, which has been proven on~\cite{LLAMA} for text generation and ~\cite{ScalableDiffusionModelsWTransformers} for image generation. We instead, show that our designed transformer architecture can also be scaled up for the animatable Gaussian avatar reconstruction.

After the feature extraction process, each point retains its unique features. Leveraging these features, we construct decoding headers $\mathcal{D}$ made up of MLPs to predict each point's Gaussian attributes, including color $c_k \in \mathbb{R}^3$, opacity $o_k \in \mathbb{R}$, per-axis scale factors $s_{k} \in \mathbb{R}$, and rotation $R_k \in SO(3)$. Since the FLAME only provides the coarse shape information of a person, without details like hairs, we also predict each point's offset to obtain a more detailed shape of the person, denoted as $O_k \in \mathbb{R}^3$. The decoding process can then be denoted as:
\begin{equation}
    \{c_k, o_k, s_k, R_k, O_k\}^{M}_{k=1} = \mathcal{D}(F_{P_{L}}),
\end{equation}
where $M$ is the total number of Gaussians. Consequently, we reconstruct the Gaussian avatar in the canonical space, which can then be animated with new expressions and rendered through a splatting process to produce the final image.

\paragraph{\textbf{FLAME Subdivision.}}
We utilize FLAME vertices as initialization of the Gaussian position and also utilize them to query features from the image to predict the Gaussian attributes. However, the original number of vertices on the FLAME template model is 5023, which is small and Gaussian avatar with such a number of points are not able to reconstruct details like hair, moustache, and teeth well. Under such observation, we propose to increase the number of points from FLAME as initialization. 
Specifically, we utilize the mesh subdivision algorithm to increase the number of mesh faces and vertices. Specifically, for each subdivision iteration, the triangle mesh is subdivided by adding a new vertex at the center of each edge and dividing each face into four new faces. We also attach blendshapes attributes to each vertex, and subdivide them by averaging the values of the attributes at the two vertices that form each edge. In this way, we increase the number of vertices and obtain the blendshapes attributes of each vertice for animation. These blendshapes attributes are also applied to each Gaussian point for the animation of the reconstructed Gaussian avatar. We can apply different numbers of mesh subdivision iterations to obtain different numbers of vertices as initialization. More points will lead to more detailed reconstruction but at a lower rendering speed. By default, we apply mesh subdivision twice to obtain $M=81,424$ vertices as initialization for $M$ Gaussian points, which is the best trade-off between reconstruction quality and rendering speed from our experiments.

\subsection{New Expression Animation}
Given the reconstructed canonical Gaussian head avatar, we introduce how to reenact them into new FLAME expressions in this section. Since our Gaussian positions are initialized from FLAME and attached with their corresponding animation attributes, we can animate the model utilizing standard vertex-based linear blend skinning (LBS) and corrective blendshapes as the FLAME model did. We first have a brief review of how the FLAME model is animated. Denote $\bar{T}$ the vertexes of the FLAME template mesh, $B_{S}(\vec{\beta})$, the shape blendshape function, $B_{P}(\vec{\theta})$ the pose blendshapes function, $B_{E}(\vec{\phi})$ the expression blendshape function, and $\mathcal{S}(\bar{T}, \textbf{J}, \vec{\theta}, \mathcal{W})$ the standard skinning function to rotate the vertexes $\bar{T}$ around joints $\textbf{J}$ and smoothed by blendweights $\mathcal{W}$, the FLAME model can be defined as:
\begin{equation}
    \label{eqn:LBS}
    F(\vec{\beta}, \vec{\theta}, \vec{\phi}) = \mathcal{S}(T_{P}(\vec{\beta}, \vec{\theta}, \vec{\phi}), \textbf{J}(\vec{\beta}), \vec{\theta}, \mathcal{W}),
\end{equation}
where
\begin{equation}
    \label{eqn:blenshape}
    T_{P}(\vec{\beta}, \vec{\theta}, \vec{\phi}) = \bar{T} + B_{S}(\vec{\beta}; \mathcal{S}) + B_{P}(\vec{\theta}; \mathcal{P}) + B_{E}(\vec{\phi}; \mathcal{E}),
\end{equation}
with $T_{P}(\vec{\beta}, \vec{\theta}, \vec{\phi})$ the template with added shape, pose, and expression offsets. Our reconstructed Gaussian avatar can be reenacted similarly by regarding the position location as the vertices of FLAME and animating it with the blendshape and LBS function to obtain the animated new Gaussian positions. The difference is that we utilize the vertices of the subdivided FLAME mesh after shape blendshapes $\bar{T} + B_{S}(\vec{\beta})$ as the initial Gaussian position and our networks predict the XYZ offset of each point to further improve the shape details in the canonical space, the function in Formula~(\ref{eqn:blenshape}) can be revised as:
% \weihaoquestion{without $\beta$ ?}
\begin{equation}
    \label{eqn:gsblendshape}
    T_{G}(\vec{\theta}, \vec{\phi}) = \bar{G} + B_{P}(\vec{\theta}; \mathcal{P}) + B_{E}(\vec{\phi}; \mathcal{E}), 
\end{equation}
where 
\begin{equation}
    \label{eqn:gsshapeblenshape}
    \bar{G} = \bar{T} + B_{S}(\vec{\beta}; \mathcal{S}) + O,
\end{equation}
with $O = \mathcal{N}(\bar{T} + B_{S}(\vec{\beta}; \mathcal{S}))$ the predicted offset from our network $\mathcal{N}$ given shape blended vertexes as input. Note that our shape blendshapes procedure denotes in Formula~(\ref{eqn:gsshapeblenshape}) only computes once and can be fixed once the network is forwarded and the canonical Gaussian avatar $\bar{G}$ is reconstructed. Meanwhile, we can also pre-compute the shape blendshapes to the joint once as $\bar{J}=\textbf{J}(\vec{\beta})$. Thereafter, we can then pass the reconstructed $\bar{G}$ into the LBS function to animate the Gaussian avatar at a fast speed during inference as:
\begin{equation}
    \label{eqn:LBSGS}
    F_{G}(\vec{\theta}, \vec{\phi}) = \mathcal{S}(T_{G}(\vec{\theta}, \vec{\phi}), \bar{J}, \vec{\theta}, \mathcal{W}).
\end{equation}
After the position of the Gaussian is reanimated, we perform splatting on the reanimated Gaussian and render the new image. Note that the animation processing in Formula~(\ref{eqn:LBSGS}) and the rendering does not have any neural network and postprocessing, which can be directly integrated into the traditional rendering pipeline developed for various devices, enabling fast deployment, which we will discuss in Sec.~\ref{sec:deploy_various_devices}.

\begin{figure}
    \centering
    \includegraphics[width=1\linewidth]{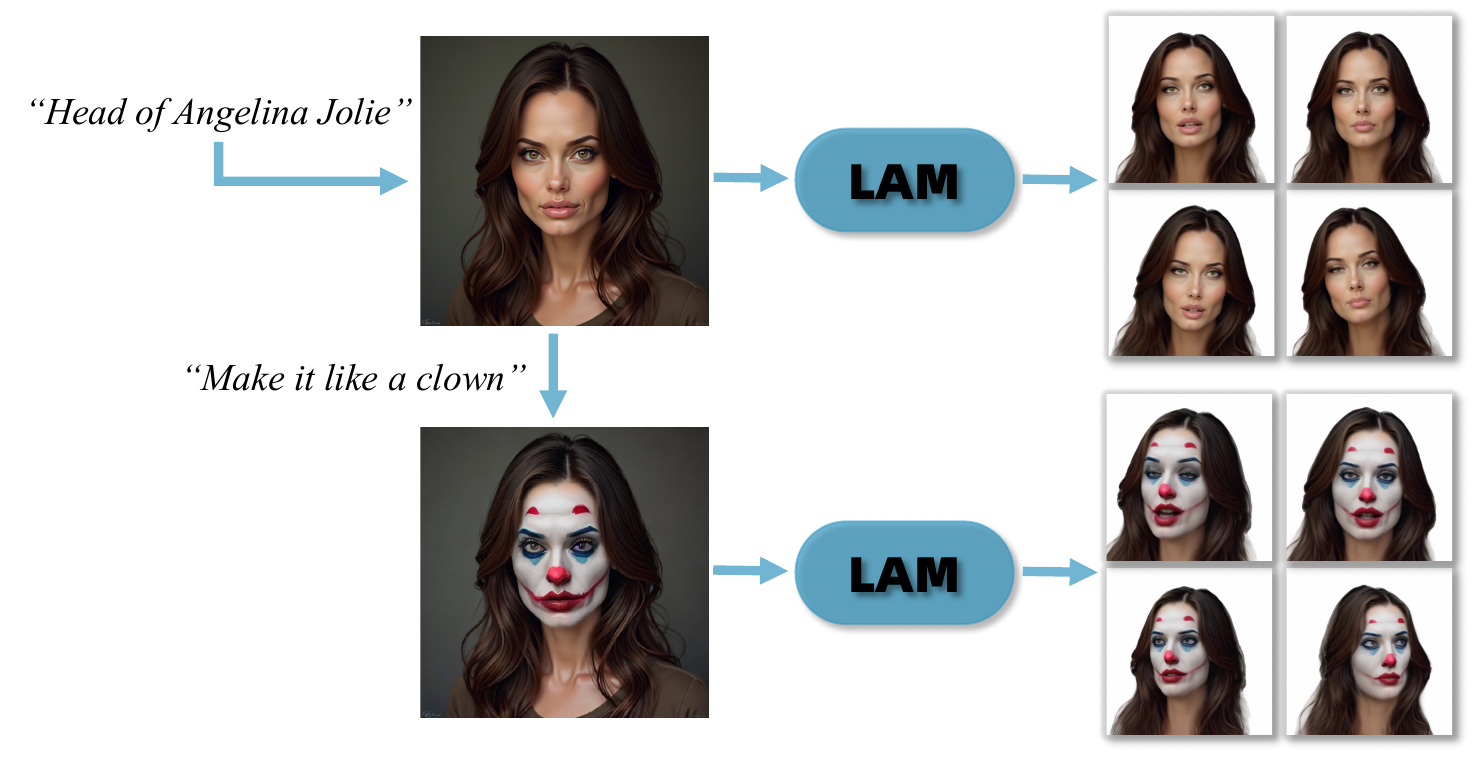}
    % \vspace{1.5in}
    \caption{Text to animatable Gaussian avatar generation and editing pipelines.}
    \label{fig:text-style-gen}
    \vspace{-0.1in}
\end{figure}

\subsection{Text to Animatable Gaussian Avatar Generation and Editing.}
Our approach takes full advantage of the shape information and animation attributes provided by the FLAME model and the robust capabilities of the DinoV2 image feature extractor, which is trained on extensive image datasets. Furthermore, after training on large-scale datasets, the scaling capability of the designed Transformer-based architecture allows our framework to adapt effectively to various input styles, including those derived from generated images. Building on this foundation, we have designed a streamlined pipeline for converting text prompts into animatable Gaussian avatars, along with an efficient style editing process, as illustrated in Fig.~\ref{fig:text-style-gen}. To generate target portrait images based on textual prompts, we employ established text-to-image generation frameworks, such as Stable Diffusion~\cite{stablediffusion}. The resulting images are then processed by our system to create animatable Gaussian avatars through a feed-forward mechanism. In addition, our framework is equipped to facilitate style editing; we can take an input image and utilize an image-to-image translation framework to modify the portrait's style, such as changing the age or transforming it into a cartoon representation. This edited image is subsequently fed into our model to produce animatable Gaussian avatars that reflect the desired styles.

\subsection{Deployment to Traditional Rendering Pipeline on Various Platforms.}
\label{sec:deploy_various_devices}
To deploy our Gaussian avatar on various computing platforms, we choose WebGL as our implementation framework, the application of which can be run on web browsers of different devices. Specifically, since the animation mainly consists of matrix operations, and the computing is highly parallelized, we pass the blenshapes and LBS information as texture to the GLSL vertex shader and use transform feedback for efficient GPU computing. For Gaussian splating, we also implement a WebGL version. In this way, we enable animation and rendering of reconstructed Gaussian avatars on different devices. Our interactive viewer is an HTML webpage with Javascript, rendered by WebGL.

\subsection{Optimization and Regularization}

During training, we randomly sample $N_f$ frames of images from the same video, selecting one as the reference image to reconstruct the canonical Gaussian avatar and others as the driving images and target images.
We supervise the rendered RGB image with the ground truth target images with a combination of $\mathcal{L}_1$ loss and perceptual loss:
\begin{equation}
    \mathcal{L}_{rgb} = \lambda_{1}\mathcal{L}_1 + \lambda_{2}\mathcal{L}_{lpips}.
\end{equation}
We also render the silhouette and supervised it with $\mathcal{L}_1$ loss denoted as $\mathcal{L}_{mask}$.

We predict the deformation offset for each point to obtain a better shape that can cover areas not modeled by FLAME, such as hair and accessories. Without constraint, the points may move freely and hurt the animation results. Therefore, we also add a regularization on the predicted offset:
\begin{equation}
    \mathcal{L}_{o} = \mathcal{L}_2(O, \epsilon),
\end{equation}
where $O$ is the predicted offset, $\epsilon$ an hyper parameters set close to 0 so that each point will not deform too much.
The total loss function is a weighted sum of the above items as:
\begin{equation}
    \mathcal{L} = \lambda_1\mathcal{L}_1 + \lambda_2\mathcal{L}_{lpips} + \lambda_3\mathcal{L}_{mask} + \lambda_4\mathcal{L}_{o}.
\end{equation}

\begin{table*}[ht]
    \centering
    \caption{Quantitative results on the VFHQ dataset. }
    \label{tab:results_vfhq}
    \vspace{-.05in}
    \begin{tabular}{l|ccccccc|ccc}
        \toprule
        & \multicolumn{7}{c|}{\textbf{Self Reenactment}} & \multicolumn{3}{c}{\textbf{Cross Reenactment}} \\
        \textbf{Method} & \textbf{PSNR$\uparrow$} & \textbf{SSIM$\uparrow$} & \textbf{LPIPS$\downarrow$} & \textbf{CSIM$\uparrow$} & \textbf{AED$\downarrow$} & \textbf{APD$\downarrow$} & \textbf{AKD$\downarrow$} & \textbf{CSIM$\uparrow$} & \textbf{AED$\downarrow$} & \textbf{APD$\downarrow$} \\
        \midrule
        StyleHeat~\citep{styleheat2022}             & 19.95      & 0.726      & 0.211      & 0.537      & 0.199      & 0.385      & 7.659      & 0.407      & 0.279      & 0.551      \\
        ROME~\citep{rome2022}                       & 19.96      & 0.786      & 0.192      & 0.701      & 0.138      & {0.186} & 4.986      & 0.530      & {0.259} & 0.277      \\
        OTAvatar~\citep{otavatar2023}               & 17.65      & 0.563      & 0.294      & 0.465      & 0.234      & 0.545      & 18.19      & 0.364      & 0.324      & 0.678      \\
        HideNeRF~\citep{hidenerf2023}               & 19.79      & 0.768      & 0.180      & 0.787      & 0.143      & 0.361      & 7.254      & 0.514      & 0.277      & 0.527      \\
        GOHA~\citep{goha2023}                       & 20.15      & 0.770      & {0.149} & 0.664      & 0.176      & {0.173} & 6.272      & 0.518      & 0.274      & {0.261} \\
        CVTHead~\citep{ma2024cvthead}               & 18.43      & 0.706      & 0.317      & 0.504      & 0.186      & 0.224      & 5.678      & 0.374      & 0.261      & 0.311      \\
        GPAvatar~\citep{gpavatar2024}               & {21.04} & {0.807} & 0.150      & 0.772      & {0.132} & 0.189      & {4.226} & 0.564      & {0.255} & 0.328      \\
        Real3DPortrait~\citep{ye2024real3d}         & 20.88      & 0.780      & 0.154      & {0.801} & 0.150      & 0.268      & 5.971      & \textbf{0.663} & 0.296      & 0.411      \\
        Portrait4D~\citep{deng2024portrait4d}       & 20.35      & 0.741      & 0.191      & 0.765      & 0.144      & 0.205      & 4.854      & 0.596      & 0.286      & {0.258} \\
        Portrait4D-v2~\citep{deng2024portrait4d2}   & {21.34} & {0.791} & {0.144} & {0.803} & {0.117} & 0.187      & {3.749} & {0.656} & 0.268      & 0.273      \\
        GAGAvatar~\cite{GAGAvatar} & 21.83 & 0.818 & 0.122 & 0.816 & 0.111 & 0.135 & 3.349 & 0.633 & 0.253 & \textbf{0.247} \\
        \midrule
        \textbf{Ours} & \textbf{22.65} & \textbf{0.829} & \textbf{0.109} & \textbf{0.822} & \textbf{0.102} & \textbf{0.134} & \textbf{2.059} & {0.651} & \textbf{0.250} & {0.356} \\
        \bottomrule
    \end{tabular}
    \vspace{-.05in}
\end{table*}

\begin{table*}[h]
\centering
\caption{Quantitative results on the HDTF dataset.}
\label{tab:results_hdtf}
    \vspace{-.05in}
\begin{tabular}{l|ccccccc|ccc}
\toprule
& \multicolumn{7}{c|}{\textbf{Self Reenactment}} & \multicolumn{3}{c}{\textbf{Cross Reenactment}} \\
\textbf{Method} & \textbf{PSNR$\uparrow$} & \textbf{SSIM$\uparrow$} & \textbf{LPIPS$\downarrow$} & \textbf{CSIM$\uparrow$} & \textbf{AED$\downarrow$} & \textbf{APD$\downarrow$} & \textbf{AKD$\downarrow$} & \textbf{CSIM$\uparrow$} & \textbf{AED$\downarrow$} & \textbf{APD$\downarrow$} \\ 
\midrule
StyleHeat~\citep{styleheat2022}             & 21.41      & 0.785      & 0.155      & 0.657      & 0.158      & 0.162      & 4.585      & 0.632      & 0.271      & 0.239      \\
        ROME~\citep{rome2022}                       & 20.51      & 0.803      & 0.145      & 0.738      & 0.133      & {0.123} & 4.763      & 0.726      & 0.268      & 0.191      \\
        OTAvatar~\citep{otavatar2023}               & 20.52      & 0.696      & 0.166      & 0.662      & 0.180      & 0.170      & 8.295      & 0.643      & 0.292      & 0.222      \\
        HideNeRF~\citep{hidenerf2023}               & 21.08      & 0.811      & 0.117      & {0.858} & 0.120      & 0.247      & 5.837      & 0.843      & 0.276      & 0.288      \\
        GOHA~\citep{goha2023}                       & 21.31      & 0.807      & 0.113      & 0.725      & 0.162      & {0.117} & 6.332      & 0.735      & 0.277      & \textbf{0.136} \\
        CVTHead~\citep{ma2024cvthead}               & 20.08      & 0.762      & 0.179      & 0.608      & 0.169      & 0.138      & 4.585      & 0.591      & {0.242} & 0.203      \\
        GPAvatar~\citep{gpavatar2024}               & {23.06} & {0.855} & {0.104} & 0.855      & {0.114} & 0.135      & {3.293} & 0.842      & 0.268      & 0.219      \\
        Real3DPortrait~\citep{ye2024real3d}         & 22.82      & 0.835      & {0.103} & 0.851      & 0.138      & 0.137      & 4.640      & \textbf{0.903} & 0.299      & 0.238      \\
        Portrait4D~\citep{deng2024portrait4d}       & 20.81      & 0.786      & 0.137      & 0.810      & 0.134      & 0.131      & 4.151      & 0.793      & 0.291      & 0.240      \\
        Portrait4D-v2~\citep{deng2024portrait4d2}   & {22.87} & {0.860} & {0.105} & {0.860} & {0.111} & {0.111} & {3.292} & {0.857} & {0.262} & {0.183} \\
        GAGAvatar~\cite{GAGAvatar}                  & 23.13 & 0.863 & 0.103 & 0.862 & 0.110 & 0.111 & 2.985 & 0.851 & 0.231 & 0.181  \\
\midrule
        Ours                                        & \textbf{23.43} & \textbf{0.873} & \textbf{0.097} & \textbf{0.865} & \textbf{0.101} & \textbf{0.093} & \textbf{1.965} & 0.849 & \textbf{0.230} & 0.229  \\
\bottomrule
\end{tabular}
\label{tab:hdtf_results}
    \vspace{-.05in}
\end{table*}

\setlength{\tabcolsep}{3pt}
\begin{table*}[h]
    \centering
    \caption{Running time of reenactment and rendering measured in FPS. All results exclude the time for avatar reconstruction and driving parameters estimation that can be calculated in advance. The results are averaged over 100 frames.}
    \label{tab:fps}
    \vspace{-.1in}
    \begin{tabular}{l|cccccccc|ccc}
        \toprule
& \multicolumn{8}{c|}{Platform with A100 GPU} & \multicolumn{3}{c}{Ours on Different Platforms} \\
        Method & StyleHeat & ROME & HideNeRF & CVTHead & Real3D & P4D-v2 & GAGavatar & Ours & Macbook (M1 Pro) & iPhone 16 & Xiaomi 14 \\
        \midrule
        FPS    & 19.82 & 11.21 & 9.73 & 18.09 & 4.55 & 9.62 & 67.12 & \textbf{280.96} & 120 & 35 & 26 \\
        \bottomrule
    \end{tabular}
    \vspace{-.01in}
\end{table*}
\setlength{\tabcolsep}{5pt}

\section{Experiments}

\subsection{Experiments Setting}
\paragraph{\textbf{Implementation Details.}}
We implement our framework with PyTorch. We froze the DINOv2 image feature extraction backbone. The Transformer architecture consists of $L=10$ layers of basic Transformer blocks with 16 attention heads and 1024 feature dimensions, and the extracted features are mapped to Gaussian attributes with one Linear layer. We use estimated FLAME as our driving 3DMM. We train the network with ADAM optimizer and cosine warm-up scheduler for 200 epochs. We set the hyper-parameters $N_{f}=8$, $\lambda_{1}=\lambda_{2}=\lambda_{3}=1$, and $\lambda_{4}=0.1$ empirically.

\paragraph{\textbf{Datasets.}}
We utilize the VFHQ dataset~\cite{vfhq} for training our model, which consists of video clips from a variety of interview scenarios. The dataset comprises 15,204 video clips with 3M frames. For each frame extracted from the videos, we detect the face region, enlarge the bounding box to crop out the region of interest, and resize the cropped images to 512×512 pixels for uniformity following GAGAvatar~\cite{GAGAvatar}. To enhance the dataset's usability, we implement tracking for camera poses and FLAME parameters as in GaussianAvatar~\cite{GaussianAvatar}. Following the methodology established by \cite{GPAvatar}, we also conducted background removal to isolate the subjects in the frames. For evaluation purposes, we employ sampled frames from the original test split of the VFHQ dataset following previous works~\cite{GPAvatar,GAGAvatar}. In our setup, the first frame of each video is designated as the source image, while the subsequent frames act as driving and target images for the reenactment process. Additionally, we conduct evaluations using the HDTF dataset~\cite{HTDF} adhering to the test split established by prior works~\cite{OTAvatar}. This evaluation includes a selection of 19 video clips, complementing our analysis and contributing to a robust assessment of our model’s performance across different datasets. 

\paragraph{\textbf{Evaluation Metrics.}}
In our evaluation of single-image animatable Gaussian head avatar generation, we focus on self and cross-identity reenactment performance metrics. For self-reenactment, when ground truth images are available, we assess the quality of the synthesized images using three quantitative measures: Peak Signal-to-Noise Ratio (PSNR), Structural Similarity Index (SSIM), and Learned Perceptual Image Patch Similarity (LPIPS). These metrics allow us to compare the synthetic images against the ground truth effectively. To evaluate identity similarity (CSIM), we compute the cosine distance of face recognition features, building on the work of~\cite{ArcFace}. For assessing expression and pose fidelity, we employ the Average Expression Distance (AED) and Average Pose Distance (APD) as determined by a 3D Morphable Model (3DMM) estimator~\cite{AEDAPD}. Additionally, we measure the Average Keypoint Distance (AKD) using a facial landmark detector as proposed by~\cite{AKD}. These metrics provide insights into the accuracy of the driving control in our animations. In cross-identity reenactment, where ground truth images are unavailable, we rely on CSIM, AED, and APD for evaluation. These metrics are in line with methodologies used in prior research~\cite{GAGAvatar}, allowing for a coherent comparison across different approaches.

\subsection{Main Results}
\begin{figure*}
  \centering
    \includegraphics[width=1.\linewidth]{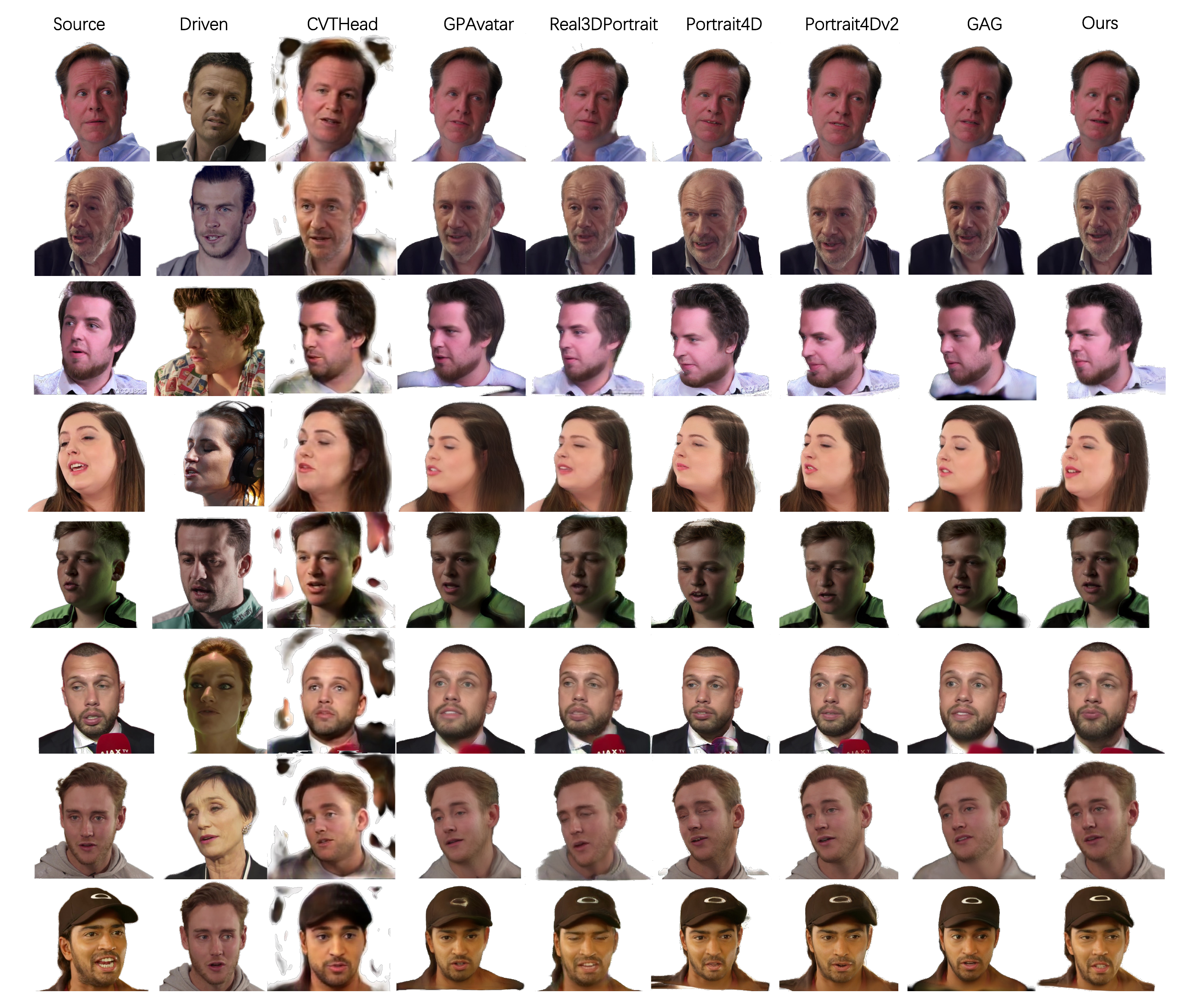}
  \caption{Visualization of cross-reenacted results on the VFHQ dataset.}
  \Description{}
  \label{fig:vis_cross_reenact}
\end{figure*}

\paragraph{\textbf{Qualitative Results.}}
Fig.~\ref{fig:vis_cross_reenact} compare the cross-reenactment results with other methods on the VFHQ dataset. Compared to previous methods, LAM achieves better reconstruction details on the textures, preserves better identity consistency, and presents more consistent expression and pose as the driven image. Note that our framework does not utilize any super-resolution and post-processing techniques and can be easily deployed on a traditional rendering pipeline for real-time rendering on various computing platforms, including mobile phones.

\paragraph{\textbf{Quantitative Results.}}
We also benchmark our framework on the VFHQ and HDTF dataset. Table~\ref{tab:results_vfhq} and Table~\ref{tab:results_hdtf} show the results on the two datasets respectively. As is shown in the table, our method achieves the best image reconstruction quality as is revealed from the PSNR, SSIM, and LPIPS metrics. Meanwhile, we preserve good identity consistency, as is shown with the CSIM metric. Furthermore, we obtain accurate expression and pose consistency with the driven image, as is revealed from the AED, APD, and AKD metrics. Notably, our framework achieves such quantitative results with extremely faster animation and rendering speeds, enabling real-time rendering on a wide range of devices.

\paragraph{\textbf{Inference Speed on Various Platforms.}} In the left side of Table~\ref{tab:fps}, we evaluate all methods on an A100 GPU. With a naive PyTorch framework and official 3D Gaussian Splatting implementation, our framework achieves extremely faster animation and rendering speed compared to existing methods. We also integrate our generated Gaussian avatar into WebGL framework and deploy them on various computing devices. As shown on the right side of Table~\ref{tab:fps}, our generated Gaussian avatar can perform real-time animation and rendering across a wide range of computing platforms, including mobile phones.

\begin{table}
    \caption{Effect of different design choices on the VFHQ dataset.}
    \vspace{-.1in}
    \label{tab:ablation}
    \small
    \begin{tabular}{l|ccc|ccc}
        \toprule
        Choices & LAM-5K & LAM-20K & LAM-80K & Tri. & PE & Paint \\
        % 5,143 & 20,426 & 81,424 \\
        \midrule
        PSNR          & 20.96  & 21.43  & \textbf{22.65}  & 21.33 & 20.96 & 20.87 \\
        FPS           & \textbf{705.63}  & 562.97  & 280.96 & 280.96 & 280.96 & 280.96 \\
        \bottomrule
    \end{tabular}
\end{table}

\subsection{More Applications}

\begin{figure*}
  % \vspace{3in}
  \centering
    \includegraphics[width=1.\linewidth]{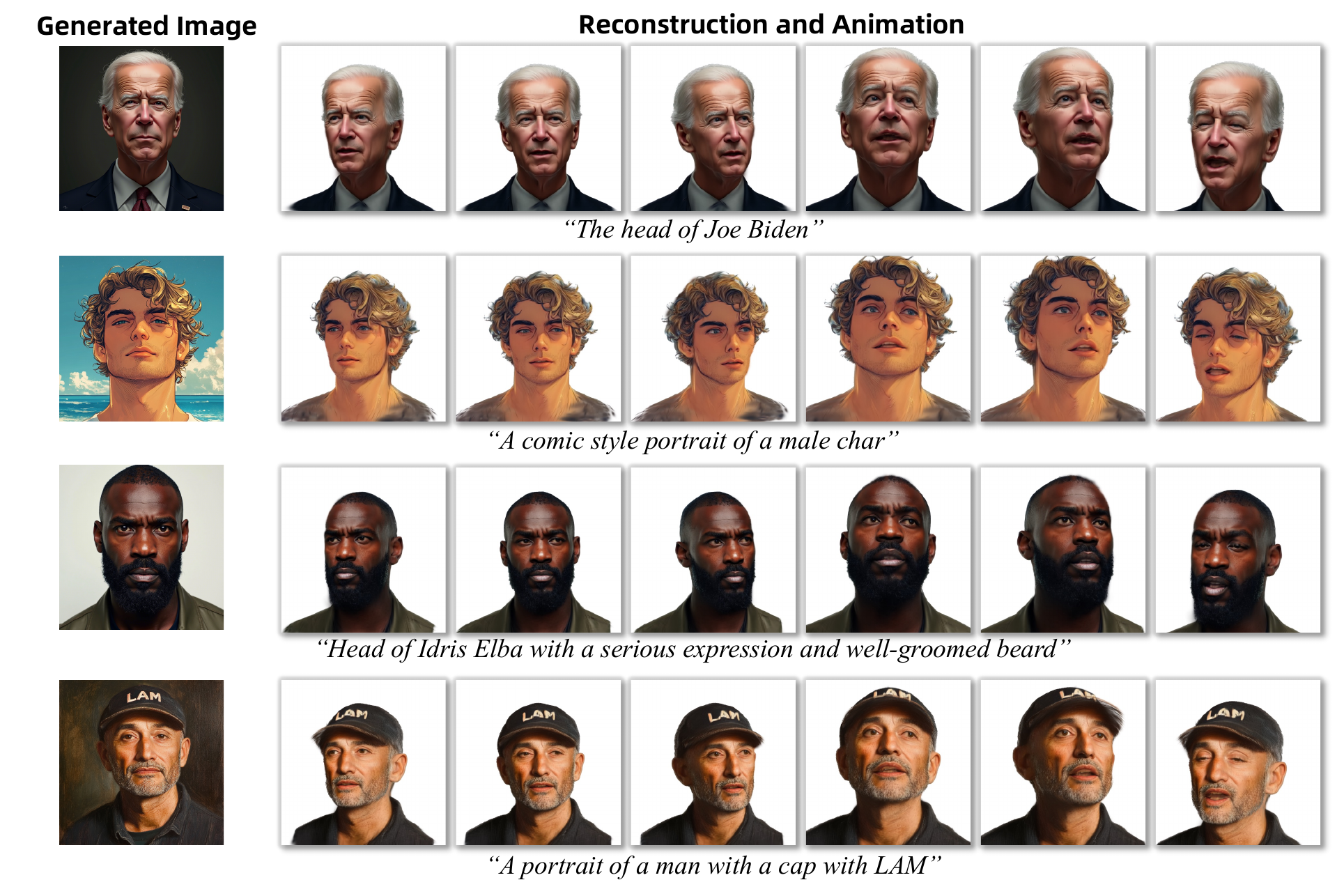}
  \caption{Visualization of cross-reenacted results on the generated image with text prompts.}
  \Description{}
  \label{fig:vis_text2image_gen}
\end{figure*}

\paragraph{\textbf{Text to Animatable Gaussian Avatar Generation.}}
Our model trained on the large-scale dataset can scale up to various image styles, including generated images from the current text-to-image dataset, this enables us to generate animatable Gaussian avatars efficiently. In Fig.~\ref{fig:vis_text2image_gen}, we reconstruct the 3D Gaussian avatar from generated images by the existing text-to-image generation pipeline and animate them with different driven images. As is shown in the figure, the reconstructed Gaussian avatar preserves the details of different styles of image, which can be animated and rendered efficiently on different platforms given different driven expressions.
 
\begin{figure*}
  % \vspace{3in}
  \centering
    \includegraphics[width=0.8\linewidth]{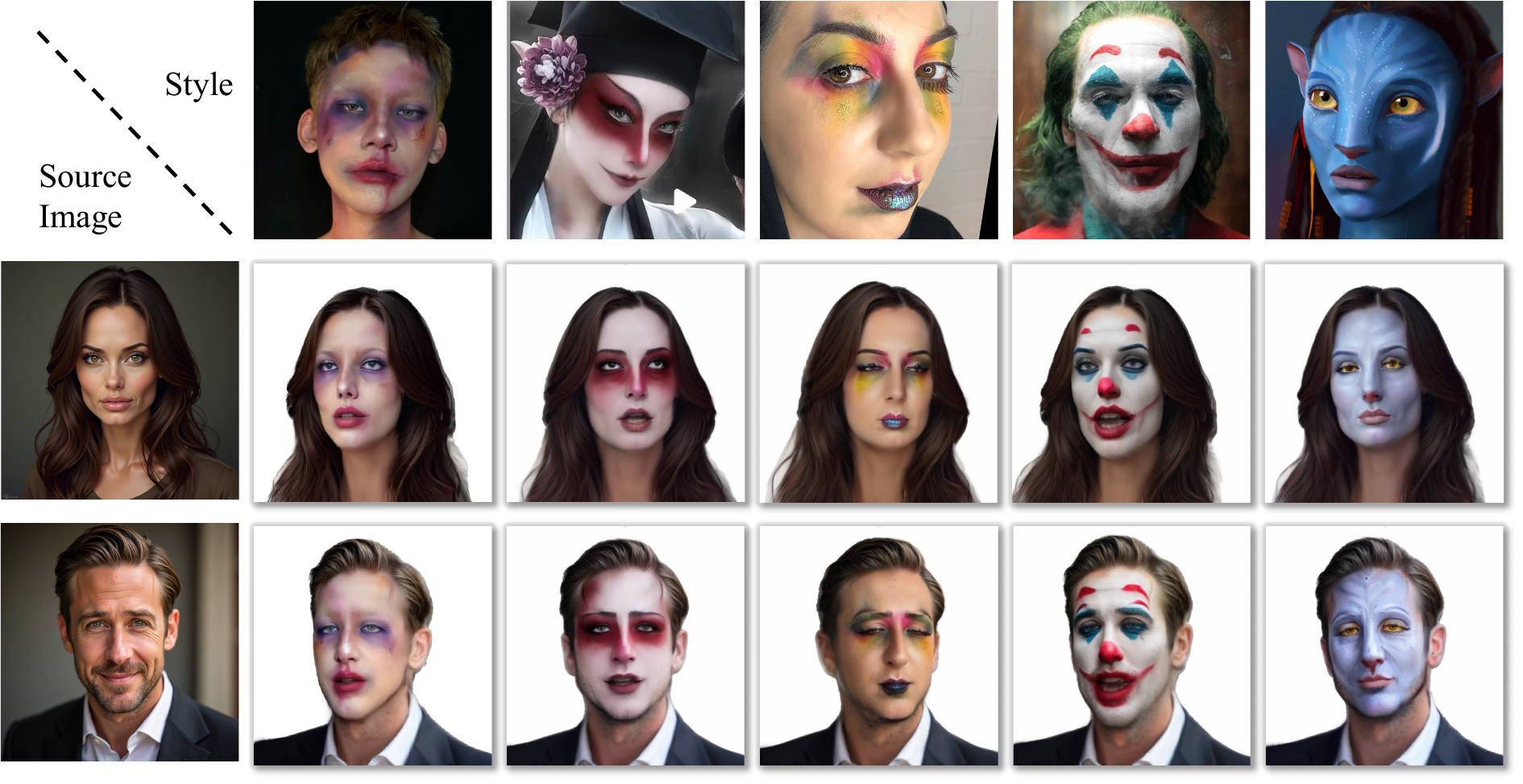}
  \caption{Visualization of cross-reenacted results on the stylized images.}
  \Description{}
  \label{fig:vis_image2image_gen}
\end{figure*}

\paragraph{\textbf{Stylize Editing of Animatable Gaussian Avatar.}}
Fig.~\ref{fig:vis_image2image_gen} shows the effective results of applying our framework for style editing of animatable Gassian avatars. Unlike previous 3D editing frameworks~\cite{IN2NICCV,FreditorECCV,GaussianEditor} that require iterative training on multi-view images for stylization, our framework can edit different styles of the 3D Gaussian avatar efficiently utilizing a 2D editing prior models to edit the avatar in the 2D image and then lift it to 3D Gaussian space, enabling efficient and user-friendly editing of 3D assets. 

\subsection{Ablation Studies}

\paragraph{\textbf{Effect of Different Number of Gaussian Points.}}
We ablate the effect of different numbers of Gaussian points (5K, 20K and 80K) in the left part of Table~\ref{tab:ablation}. We can see that more Gaussian points obtained better reconstruction quality as they can describe more details like hair and mustache, while results in slower inference speed. We find that about 80 thousand points is the best trade-off.

\paragraph{\textbf{Effect of Query Representation.}}
We also compare widely used triplane-based query representation with our point-based query. Specifically, we replace the learnable query point features $F_P$ with flattened learnable triplane features for cross-attention with the image feature. Each sampled point then fetch its corresponding features from the triplane by projection for Gaussian attributes decoding. As shown in Table~\ref{tab:ablation}, with the same number of points, our point-based framework (LAM-80K) gets better results compared with triplane-based (Tri.).  

\paragraph{\textbf{Effect of Canonical Space Generation.}} We compare our canonical space generation with directly generating Gaussian avatar in the reference expression and pose (PE in Table~\ref{tab:ablation}) and then reenact it to the driven pose and expression. The canonical space simplifies the problem and gets better results. We use the same number of points (80K) for fair comparisons.

\paragraph{\textbf{Effect of Cross-Attention on Image Features.}} Compared with only applying cross-attention on the painted image feature as in~\cite{transhuman} (denoted as Paint in Table~\ref{tab:ablation}), directly performing cross-attention on the image features gets quite better results. We use the same number of points (80K) for fair comparisons.
\section{Conclusions and Limitations}

In this work, we present a novel large avatar model for one-shot animatable Gaussian head generation. The core of our framework is the canonical Gaussian avatar generation Transformer. We utilize point-cloud representation to fully leverage the prior shape information resides in FLAME; build stacked cross-attention modules on multi-scale image features for better texture and shape reconstruction; and generate the Gaussian avatar in the unified canonical space with the same expression and pose to mitigate reconstruction complexcities.  Our framework can generate Gaussian avatars that can be seamlessly integrated into the existing rendering pipeline for real-time animation and rendering on a wide range of platforms, including mobile phones. Moreover, we introduce an efficient pipeline for text to animatable Gaussian avatar generation and a user-friendly pipeline for Gaussian avatar style editing given a single image. 

\paragraph{Limitations.} We utilize FLAME parameters to animate our reconstructed 3D Gaussian avatar. It fails to reproduce expressions that FLAME cannot model. For example, LAM is not able to model the tongue movement since FLAME does not model the tongue blend-shapes. Since we eliminate the 2D post-processing network for efficient animation and rendering, some expression-dependent details like dynamic wrinkles cannot be fully modeled. The limitation of FLAME expressiveness and the challenge of single input image also limits the expression neutralization capability. Also, since we utilize algorithms to estimate FLAME from video, the inaccurate estimated FLAME parameters will infect the results. 

\clearpage
\bibliographystyle{ACM-Reference-Format}
\bibliography{reference}

\clearpage

\clearpage
\appendix
\section{More results}
\begin{figure*}
  \centering
    \includegraphics[width=1.\linewidth]{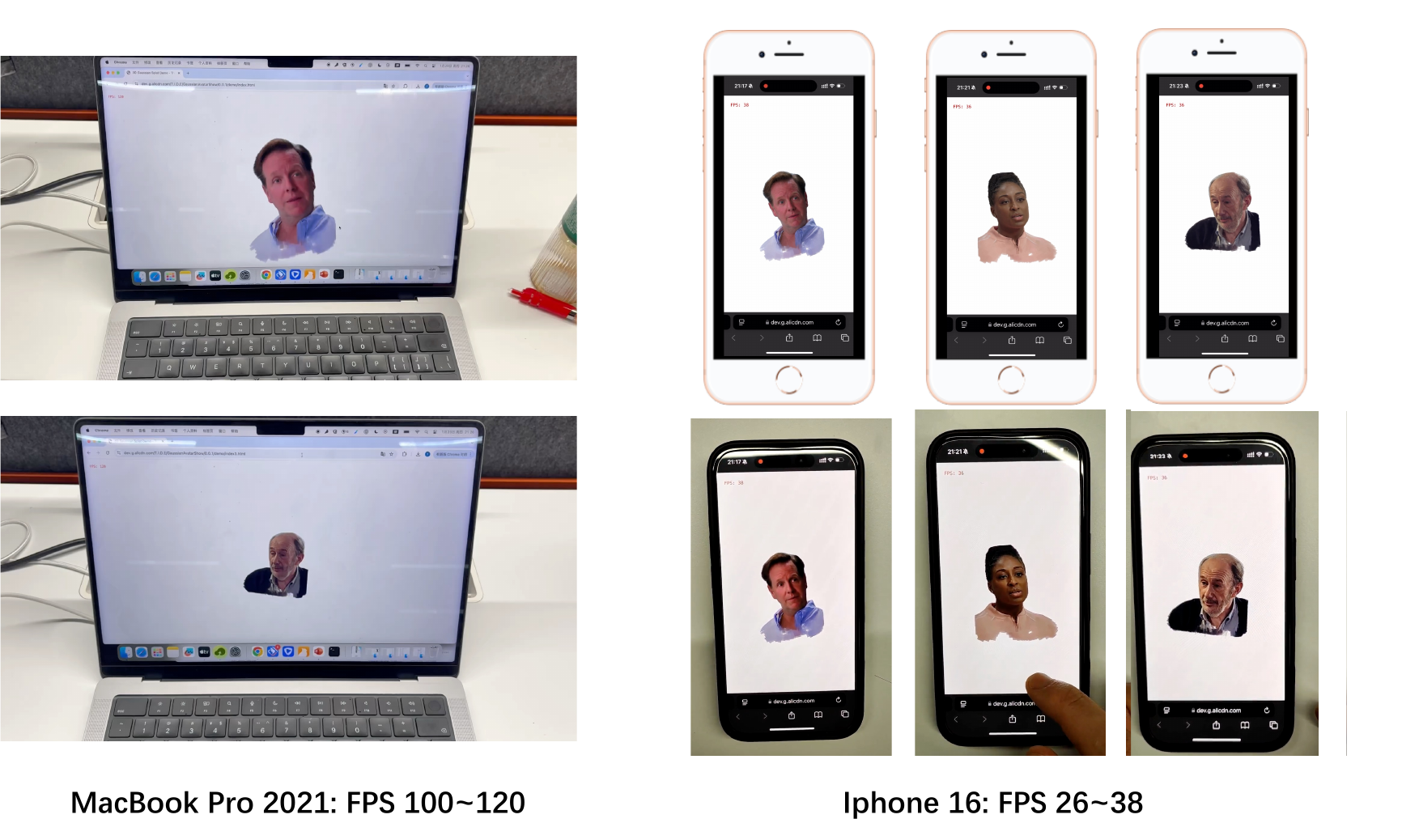}
    \vspace{-8mm}
  \caption{Visualization of the reenactment and rendering of our reconstructed Gaussian avatar on various computing platforms, including mobile phones.}
  \Description{}
  \label{fig:various_platforms}
\end{figure*}

\begin{figure*}
  % \vspace{3in}
  \centering
    \includegraphics[width=1.\linewidth]{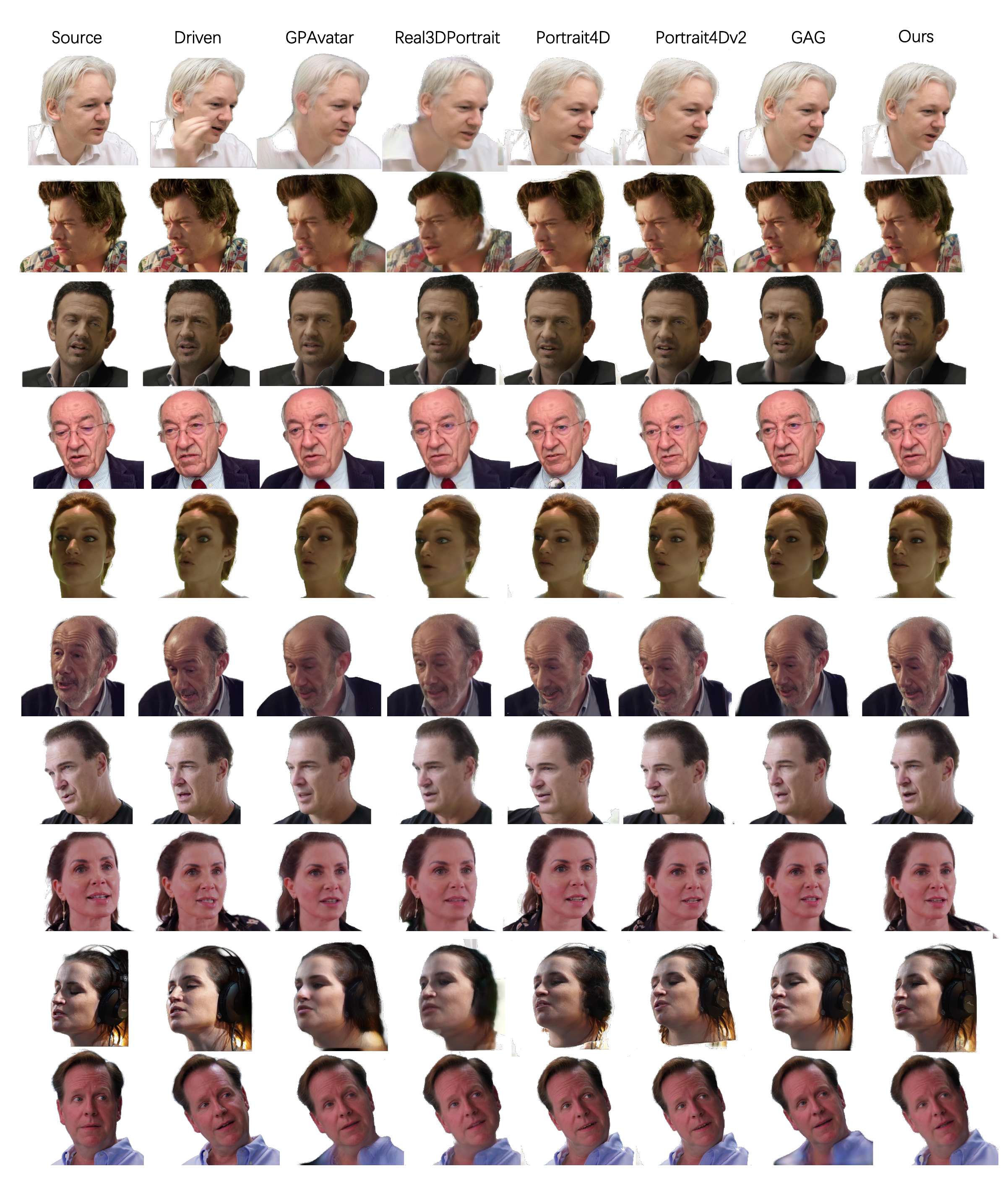}
    \vspace{-8mm}
  \caption{Visualization of self-reenacted results on the VFHQ dataset.}
  \Description{}
  \label{fig:vis_self_reenact}
\end{figure*}

We show the real-time reenactment and rendering of our reconstructed Gaussian avatars on various platforms, including mobile phones in Fig.~\ref{fig:various_platforms}. Watching our supplementary videos is strongly recommended.
In Fig.~\ref{fig:vis_self_reenact}, we show the self-reenacted results on the VFHQ dataset.

\end{document}